\documentclass[sigplan,screen]{acmart}

\AtBeginDocument{%
  \providecommand\BibTeX{{%
    \normalfont B\kern-0.5em{\scshape i\kern-0.25em b}\kern-0.8em\TeX}}}

\usepackage{microtype}
\usepackage{graphicx}
\usepackage{subfigure}
\usepackage{booktabs}
\usepackage{comment}
\usepackage{amsfonts}
\usepackage{amsmath}
\usepackage{hyperref}
\usepackage{dblfloatfix}
\usepackage{bbm}
\usepackage{fancyhdr}

\usepackage{array,booktabs}
\newcolumntype{P}[1]{>{\centering\arraybackslash}p{#1}}

\usepackage{listings}
\usepackage{color}

\usepackage{pgfplots,pgfplotstable}
\DeclareUnicodeCharacter{2212}{−}
\usepgfplotslibrary{groupplots,dateplot}
\usetikzlibrary{patterns,shapes.arrows}
\pgfplotsset{compat=newest}

\begin{document}

\title{Towards Fine-grained Visual Representations by Combining Contrastive Learning with Image Reconstruction and Attention-weighted Pooling}

\author{Jonas Dippel}
\email{jonas.dippel@bayer.com}
\affiliation{%
  \institution{Bayer AG}
  \city{Berlin}
  \country{Germany}
}

\author{Steffen Vogler}
\email{steffen.vogler@bayer.com}
\affiliation{%
  \institution{Bayer AG}
  \city{Berlin}
  \country{Germany}
}

\author{Johannes H\"ohne}
\email{johannes.hoehne@bayer.com}
\affiliation{%
  \institution{Bayer AG}
  \city{Berlin}
  \country{Germany}
}

\fancypagestyle{firstpagestyle}{%
  \fancyhf{}%
  \fancyhead[L]{accepted at ICML 2021 Workshop: Self-Supervised Learning for Reasoning and Perception}
}

\setcopyright{none}
\settopmatter{printacmref=false} 
\renewcommand\footnotetextcopyrightpermission[1]{} 

\begin{abstract}
  This paper presents Contrastive Reconstruction, ConRec - a self-supervised learning algorithm that obtains image representations by jointly optimizing a contrastive and a self-reconstruction loss. We showcase that state-of-the-art contrastive learning methods (e.g.~SimCLR) have shortcomings to capture fine-grained visual features in their representations. ConRec extends the SimCLR framework by adding (1) a self-reconstruction task and (2) an attention mechanism within the contrastive learning task. This is accomplished by applying a simple encoder-decoder architecture with two heads. 
We show that both extensions contribute towards an improved vector representation for images with fine-grained visual features. Combining those concepts, ConRec outperforms SimCLR and SimCLR with Attention-Pooling on fine-grained classification datasets.
\end{abstract}

\maketitle
\pagestyle{empty}

\begin{figure*}[th]
\vskip 0.2in
\begin{center}
\centerline{\includegraphics[width=\linewidth, trim={0cm 2.5cm 0 1cm}]{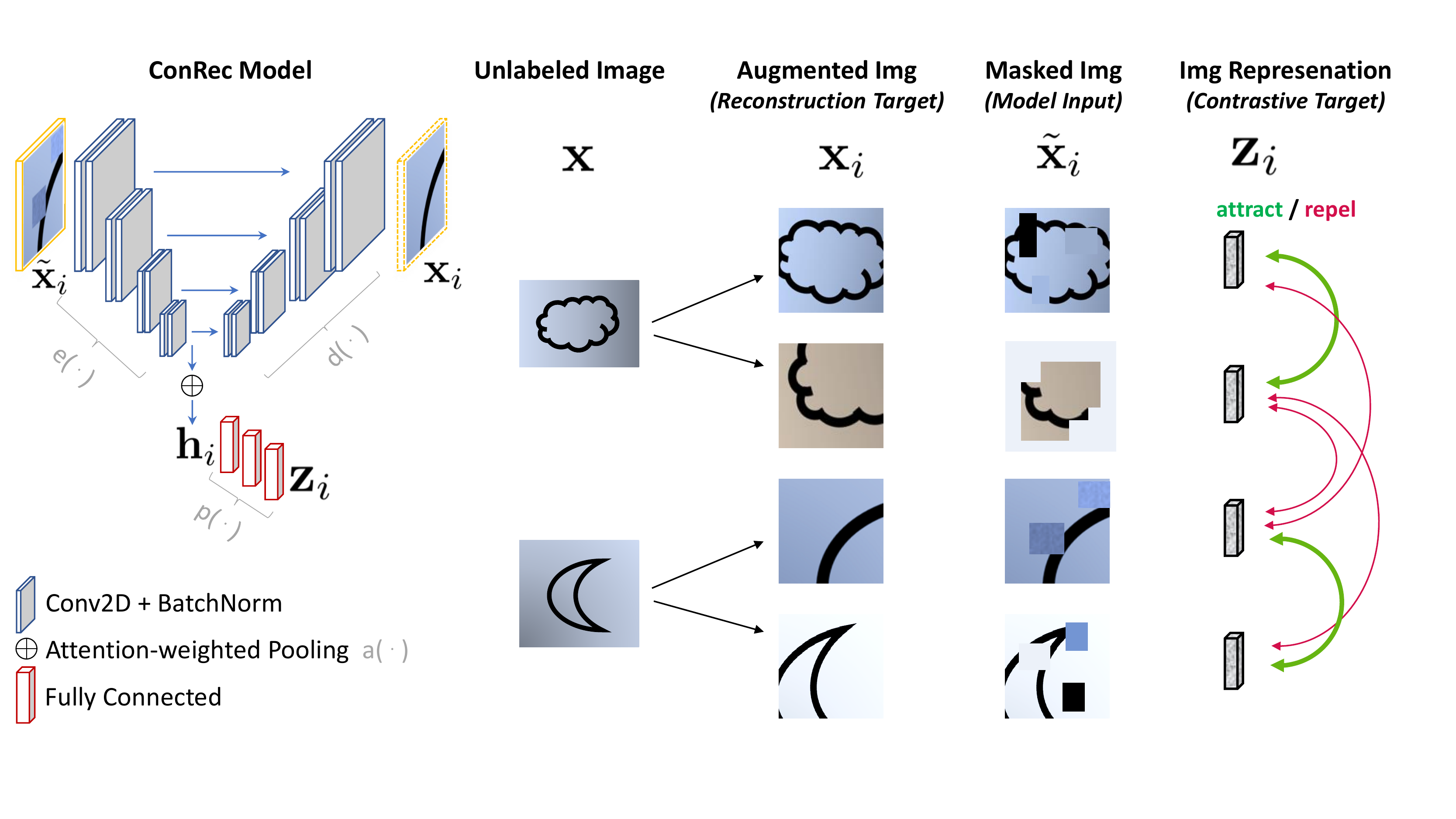}}

\caption{Learning Framework for \underline{Con}trastive \underline{Rec}onstruction -- ConRec.The ConRec model consists of a fully convolutional encoder-decoder architecture with skip connections as well as a projection head comprising fully connected layers. 
The model receives a masked image $\tilde{\mathbf{x}}_i$ and outputs the unmasked reconstruction target $\mathbf{x}_i$ as well as the contrastive image representation vector $\mathbf{z}_i$. For a batch of $N$ unlabeled images, the model receives $2N$ masked-augmented images and simultaneously performs a reconstruction task and a contrastive task. Within the contrastive task, representations $\mathbf{z}_i$ and $\mathbf{z}_j$ are optimized to be similar (dissimilar), if the inputs $i$ and $j$ arise from the same (from a distinct) original image. }
\label{fig:model}
\end{center}
\vskip -0.2in
\end{figure*}

\section{Introduction}
\label{intro}
Recent advances in the field of self-supervised learning have shown exciting progress. For the first time, \citet{he2020momentum} have reported that the concept of contrastive learning applied in unsupervised pretraining can outperform supervised pretraining used in several computer vision tasks. Additional corroborating reports \cite{sohn2020learning, misra2020self, tian2020rethinking} followed, which confirmed the notion that contrastive learning indeed leads to a representation that is well suited for challenges such as few-shot classification or domain transfer. The combination of unsupervised pretraining with supervised fine-tuning is especially attractive for many real-world application because it provides a solution to the common issue of shortage of labeled data \cite{honari2018improving,wolf2020inpainting, cao2019cross, dehaene2020self}. Large amounts of unlabeled, task-agnostic data could be leveraged to create a representation that get subjected to supervised fine-tuning using fewer labeled data \cite{chen2020big}. 

Whether representations gained with contrastive learning are suitable for fine-grained classification still needs to be studied as this kind of tasks poses a harder challenge due to subtlety of required attributes. Fine-grained classification tasks are characterized by intra-class variations being potentially greater than the inter-class variations \cite{wei2019deep}. In practice, a high degree of similarity among categories can be observed and large numbers of attributes are required to effectively model these subtle differences. Moreover, only the combination of multiple sub-regions within an image might be predictive for the class \cite{fu2017look}. In contrast, the entire image can be subject to confounding effects including different viewing angles, lighting, or partial occlusion \cite{wei2019deep, haney2020fine}. Therefore, a fine-grained classifier needs to be able to interpret fine scale features in presence of coarse scale perturbations (notably, these coarse perturbations could hold predictive value themselves). Therefore, such tasks pose a challenge to models which utilize representations derived via unsupervised pretraining.

We exemplified these shortcomings by showing that the image representations obtained through SimCLR \cite{chen2020simple, chen2020big} lead to inaccurate model performance in a carefully designed synthetic dataset for fine-grained classification.

Further, we found that the addition of an attention pooling mechanism and a self-reconstruction loss successfully addresses this limitation. Moreover, we present a novel approach that utilizes a combination of contrastive learning and a self-reconstruction task. The proposed learning paradigm yields improved performance in multiple fine-grained classification tasks, which confirms that the combination of a contrastive loss and a self-reconstruction loss leads to representations of sufficiently high granularity.

We hope this provides further understanding to future efforts to build representation that can be applied across a wide range of domains and tasks.

\section{Method}
\label{method}
The core concepts of ConRec are depicted in Figure \ref{fig:model}. Starting with N unlabeled images $\mathbf{x}$, we generate 2 augmented images for each input image, which yields 2N augmented images $\mathbf{x_i}, i\in \{1..2N\}$ that serve as reconstruction targets.
The ConRec model receives an artificially masked image $\mathbf{\tilde{x}_i}$ with the task to reconstruct $\mathbf{x_i}$. For each input $\mathbf{\tilde{x}_i}$, ConRec also outputs contrastive representations $\mathbf{z_i}$ which are optimized to be similar (dissimilar), if the inputs $i$ and $j$ originate from the same (from a distinct) unlabeled image.

\subsection{Model Architecture}
The ConRec model is a deep neural network with one input and two outputs. The model architecture can be divided into four components: encoder $e(\cdot)$, decoder $d(\cdot)$, attention-weighted pooling $a(\cdot)$ and projection head $p(\cdot)$ as shown in Figure \ref{fig:model}.
We chose the U-net \cite{ronneberger2015u} as the main backbone for the encoder and decoder for ConRec. 
In the training process, the model receives a masked image $\tilde{x}_i$ and outputs the reconstructed image $x_i = d(e(\tilde{x}_i))$ as well as the contrastive vector representation $\mathbf{z}_i = p(a(e(\tilde{x}_i)))$.
The training loss is composed of two parts: the contrastive loss $\mathcal{\mathbf{L}}_c$ and the reconstruction loss $\mathcal{\mathbf{L}}_r$.
\begin{equation}
    \mathcal{\mathbf{L}}_{ConRec} = \mathcal{\mathbf{L}}_c + \alpha * \mathcal{\mathbf{L}}_r
\end{equation}

\noindent
Reconstruction loss $\mathcal{\mathbf{L}}_r$ assesses the reconstruction quality through the pixel-wise mean squared error between the predicted reconstruction and the ground truth image.

\begin{equation}
\mathcal{\mathbf{L}}_r = \frac{1}{N} \sum_i^N (d(e(\tilde{x}_i)) - x_i)^2 .
\end{equation}

\noindent
Following the SimCLR framework, we use the normalized temperature-scaled cross entropy loss (\textit{NT-Xent}) as contrastive loss $\mathcal{\mathbf{L}}_c$ with

\begin{equation}
\mathcal{\mathbf{L}}_c^{(i,j)} = -\log\frac{\exp\left(\text{sim}\left(\mathbf{z}_{i}, \mathbf{z}_{j}\right)/\tau\right)}{\sum^{2N}_{k=1} \mathbbm{1}_{[k\neq{i}]} \exp\left(\text{sim}\left(\mathbf{z}_{i}, \mathbf{z}_{k}\right)/\tau\right)}
\end{equation}

\noindent
where $\tau$ depicts the temperature parameter. We refer to Algorithm 1 in  \citet{chen2020simple} for a precise description of loss term $\mathcal{\mathbf{L}}_c$.

After the training is completed, we discard the projection head and the decoder and we only use the encoder $e(\cdot)$ and the attention pooling $a(\cdot)$ to generate image representations with $\mathbf{h}_i = a(e(x_i))$.

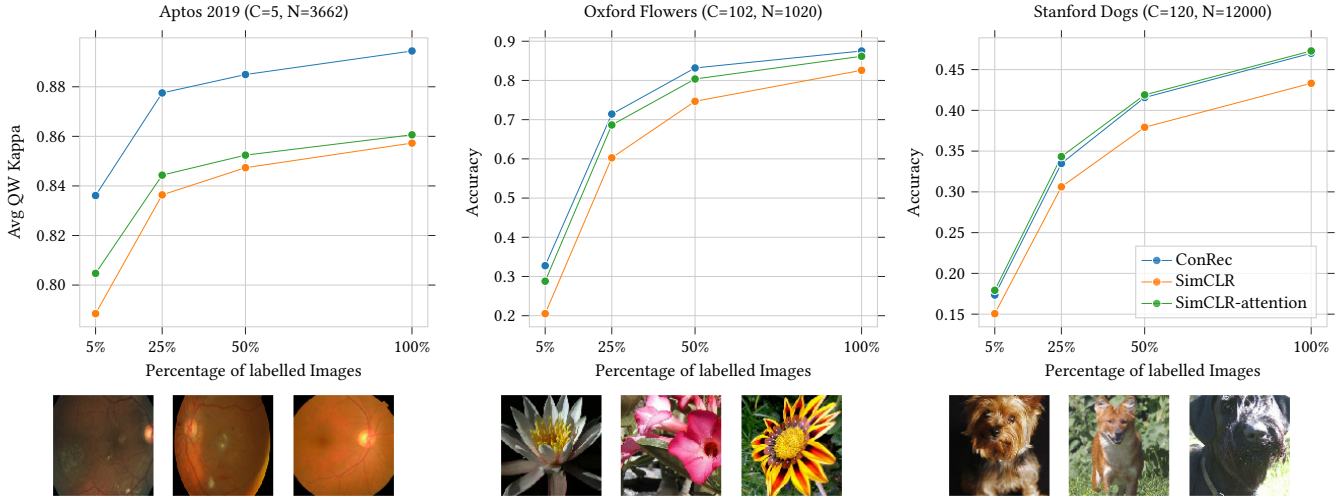
\begin{figure*}[th]
\begin{center}
\resizebox{\textwidth}{!}{
\begin{tikzpicture}

\definecolor{color0}{rgb}{0.12156862745098,0.466666666666667,0.705882352941177}
\definecolor{color1}{rgb}{1,0.498039215686275,0.0549019607843137}
\definecolor{color2}{rgb}{0.172549019607843,0.627450980392157,0.172549019607843}

\begin{groupplot}[group style={group size=3 by 1,horizontal sep=2cm, vertical sep=0cm}]
\nextgroupplot[
axis line style={white!80!black},
legend style={fill opacity=0.8, draw opacity=1, text opacity=1, draw=none},
tick align=outside,
title={Aptos 2019 (C=5, N=3662)},
x grid style={white!80!black},
xlabel={Percentage of labelled Images},
xmajorgrids,
xmin=0.0025, xmax=1.0475,
xtick style={color=white!15!black},
xtick={0.05,0.25,0.5,1},
xticklabels={5\%,25\%,50\%,100\%},
y grid style={white!80!black},
ylabel={Avg QW Kappa},
ymajorgrids,
ymin=0.783179891109467, ymax=0.899732100963593,
ytick style={color=white!15!black},
ytick={0.78,0.8,0.82,0.84,0.86,0.88,0.9},
yticklabels={\(\displaystyle 0.78\),\(\displaystyle 0.80\),\(\displaystyle 0.82\),\(\displaystyle 0.84\),\(\displaystyle 0.86\),\(\displaystyle 0.88\),\(\displaystyle 0.90\)}
]
\addplot [line width=0.48pt, color0, mark=*, mark size=2.4, mark options={solid,draw=white}, forget plot]
table {%
0.0499999523162842 0.836127519607544
0.25 0.877513408660889
0.5 0.884933471679688
1 0.894434213638306
};
\addplot [line width=0.48pt, color1, mark=*, mark size=2.4, mark options={solid,draw=white}, forget plot]
table {%
0.0499999523162842 0.788477659225464
0.25 0.836392402648926
0.5 0.84735107421875
1 0.857280850410461
};
\addplot [line width=0.48pt, color2, mark=*, mark size=2.4, mark options={solid,draw=white}, forget plot]
table {%
0.0499999523162842 0.804735541343689
0.25 0.844353437423706
0.5 0.852415084838867
1 0.860621929168701
};
\nextgroupplot[
axis line style={white!80!black},
legend style={fill opacity=0.8, draw opacity=1, text opacity=1, draw=none},
tick align=outside,
title={Oxford Flowers (C=102, N=1020)},
x grid style={white!80!black},
xlabel={Percentage of labelled Images},
xmajorgrids,
xmin=0.0025, xmax=1.0475,
xtick style={color=white!15!black},
xtick={0.05,0.25,0.5,1},
xticklabels={5\%,25\%,50\%,100\%},
y grid style={white!80!black},
ylabel={Accuracy},
ymajorgrids,
ymin=0.171743372906164, ymax=0.908594893478614,
ytick style={color=white!15!black},
ytick={0.1,0.2,0.3,0.4,0.5,0.6,0.7,0.8,0.9,1},
yticklabels={\(\displaystyle 0.1\),\(\displaystyle 0.2\),\(\displaystyle 0.3\),\(\displaystyle 0.4\),\(\displaystyle 0.5\),\(\displaystyle 0.6\),\(\displaystyle 0.7\),\(\displaystyle 0.8\),\(\displaystyle 0.9\),\(\displaystyle 1.0\)}
]
\addplot [line width=0.48pt, color0, mark=*, mark size=2.4, mark options={solid,draw=white}, forget plot]
table {%
0.0499999523162842 0.327533006668091
0.25 0.71426248550415
0.5 0.831842660903931
1 0.875101566314697
};
\addplot [line width=0.48pt, color1, mark=*, mark size=2.4, mark options={solid,draw=white}, forget plot]
table {%
0.0499999523162842 0.205236673355103
0.25 0.602699637413025
0.5 0.746788024902344
1 0.825825333595276
};
\addplot [line width=0.48pt, color2, mark=*, mark size=2.4, mark options={solid,draw=white}, forget plot]
table {%
0.0499999523162842 0.28801429271698
0.25 0.686453104019165
0.5 0.803707838058472
1 0.861603498458862
};

\nextgroupplot[
axis line style={white!80!black},
legend cell align={left},
legend style={fill opacity=0.8, draw opacity=1, text opacity=1, at={(0.97,0.03)}, anchor=south east, draw=white!80!black},
tick align=outside,
title={Stanford Dogs (C=120, N=12000)},
x grid style={white!80!black},
xlabel={Percentage of labelled Images},
xmajorgrids,
xmin=0.0025, xmax=1.0475,
xtick style={color=white!15!black},
xtick={0.05,0.25,0.5,1},
xticklabels={5\%,25\%,50\%,100\%},
y grid style={white!80!black},
ylabel={Accuracy},
ymajorgrids,
ymin=0.134469696969697, ymax=0.488956876456877,
ytick style={color=white!15!black},
ytick={0.1,0.15,0.2,0.25,0.3,0.35,0.4,0.45,0.5},
yticklabels={\(\displaystyle 0.10\),\(\displaystyle 0.15\),\(\displaystyle 0.20\),\(\displaystyle 0.25\),\(\displaystyle 0.30\),\(\displaystyle 0.35\),\(\displaystyle 0.40\),\(\displaystyle 0.45\),\(\displaystyle 0.50\)},
]
\addplot [line width=0.48pt, color0, mark=*, mark size=2.4, mark options={solid,draw=white}]
table {%
0.0499999523162842 0.173426628112793
0.25 0.334848523139954
0.5 0.41573429107666
1 0.470046639442444
};
\addplot [line width=0.48pt, color1, mark=*, mark size=2.4, mark options={solid,draw=white}]
table {%
0.0499999523162842 0.150582790374756
0.25 0.306177139282227
0.5 0.379254102706909
1 0.43321681022644
};
\addplot [line width=0.48pt, color2, mark=*, mark size=2.4, mark options={solid,draw=white}]
table {%
0.0499999523162842 0.179254055023193
0.25 0.343240022659302
0.5 0.418997645378113
1 0.472843885421753
};
\legend{ConRec, SimCLR, SimCLR-attention};
\end{groupplot}
\end{tikzpicture}}
\begin{minipage}{.33\textwidth}
        \centering
        \includegraphics[width=.8\textwidth]{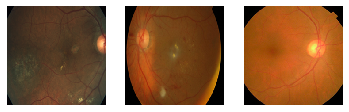}
\end{minipage}
\begin{minipage}{.33\textwidth}
    \centering
    \includegraphics[width=.8\textwidth]{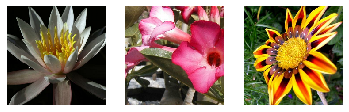}
\end{minipage}
\begin{minipage}{.33\textwidth}
    \centering
    \includegraphics[width=.8\textwidth]{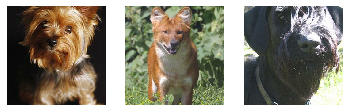}
\end{minipage}
\caption{Model accuracies for various amounts of labels. Each representation model was trained on all available training data and a linear classifier was trained on a subset (1\% to 100\%) of the labeled representations. All metrics were computed on the same fixed test data. This experiment was conducted for three datasets with different number of classes $C$ and number of samples $N$. The top row depicts the validation metric of the linear model. Sample images from the dataset are shown below.}
\label{fig:benchmark}
\end{center}
\end{figure*}

\subsection{Augmentation}
\begin{figure}[h]
\begin{minipage}{.25\linewidth}
    \centering
    Input
\end{minipage}
\begin{minipage}{.25\linewidth}
    \centering
    Prediction
\end{minipage}
\begin{minipage}{.25\linewidth}
    \centering
    Target
\end{minipage}
\begin{center}
\centerline{\includegraphics[width=0.8\linewidth]{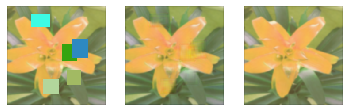}}
\centerline{\includegraphics[width=0.8\linewidth]{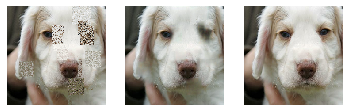}}
\centerline{\includegraphics[width=0.8\linewidth]{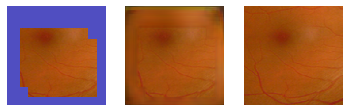}}
\caption{Augmented images and respective reconstruction predictions by our ConRec model. The images were selected from the Oxford Flowers, Stanford Dogs and Aptos 2019 dataset, ordered top to bottom. In-painting is shown in the first row, the second row shows local-pixel-shuffling and the last one image-out-painting.}
\label{fig:augmentations}
\end{center}
\end{figure}

Two types of augmentations are applied to the image. First, we apply the augmentations from SimCLR \cite{chen2020simple} including cropping, color jitter and blurring. Then, we apply augmentations that have to be reversed by the reconstruction task as described in \cite{ zhou2019models}. These augmentations include in-painting, where a randomly sized rectangle is placed on the image, out-painting where the outer parts of the image are masked and local-pixel shuffling where the pixels in a region of the image are shuffled. In the case of single channel images, we also apply non-linear transformations (see \cite{zhou2019models}). The unmasked image is then used as a target for the reconstruction task. Figure \ref{fig:augmentations} shows augmentation examples and the respective reconstruction predictions by our model.

\subsection{Attention-weighted Pooling}
SimCLR \cite{chen2020simple} uses global average pooling at the output of the encoder to reduce the dimensionality before the dense layers in the projection head. This pooling operation discards some local features in the output activation map, which may carry relevant fine-grained information.

We introduce an attention weighted pooling mechanism that aggregates the spatial content of the final feature map of the encoder in a parametric manner. This spatial weighting yields an improved image representation $\mathbf{h}$ which is then fed into the projection head. 

Suppose, we have an encoder output activation map $\phi \in \mathbb{R}^{w \times h \times f}$. First, we compute an attention weight $a_{ij}$ for each location $(i, j) \in \mathbb{N}^{w \times h}$. This is done by applying three consecutive convolution blocks (convolution, batch norm, ReLU activation) with decreasing number of filters and then one final convolution with one filter (sigmoid activation). The weights and biases of the four convolutions are trainable with the end-to-end model. 
By replicating $a_{ij}$ across dimension $f$, we obtain a matrix $A_{\phi} \in \mathbb{R}^{w \times h \times f}$. The final pooling output $a(\phi) \in \mathbb{R}^{f}$ is then computed as following:

\begin{equation}
a(x) = g(A_{\phi} \odot \phi) \odot \frac{1}{g(A_{\phi})}
\end{equation}

where $g$ denotes standard 2D global average pooling. Note that the attention-weighted pooling is only applied in the contrastive head. In analogy to our approach, \citet{radford2learning} also recently proposed a self-attention pooling mechanism as part of their CLIP model.

\subsection{Implementation Details}
In contrast to the large mini batches from SimCLR, we use a batch size of 16 resulting in a true batch size of 32 and perform model training on a single Volta V100 GPU with 16GB memory. Given the small batch size, the convergence time of SimCLR is increased.  Furthermore, compared to SimCLR, ConRec has an additional decoder which yields an increased model size and an increased training time by factor 2. For the results in Section \ref{sec:benchmark}, pretraining of the ConRec U-net model needed 270k - 900k training steps (depending on the dataset size) which results in infrastructure costs of 156 - 506\$ \footnote{on a AWS p3 instance with costs of 4,22\$ per hour} per model training. Our implementation is available on Github \footnote{https://github.com/bayer-science-for-a-better-life/contrastive-reconstruction}.
We reuse the parameters for pretraining on ImageNet. In detail, we use the LARS Optimizer with a constant learning rate of $0.3$, weight decay of $10^{-4}$ and color jitter strength of $1.0$. Furthermore, we use a temperature value of $0.5$ which yielded better results than a temperature value of $0.1$ for our datasets. To achieve the same magnitude in the loss terms, we use a factor of $\alpha=100$ for the reconstruction loss.

\subsection{Evaluation Protocol}
\label{sec:evaluation-protocol}

To separately show the effect of our attention mechanism and the reconstruction task, we compare the performance of ConRec with the SimCLR framework implementation \cite{chen2020simple} using standard global average pooling (SimCLR) and with the attention-weighted pooling mechanism (SimCLR-attention). For each method and dataset, we pretrain the model on the training set. After pretraining, we compute the representations of each image in the training and test set and use center-cropping if the images are not of the same size. For evaluation, a logistic regression classifier is trained on the representations of the training set and then evaluated on the representations of the test set. During pretraining, we evaluate our model every 20 epochs and choose the best performing model for final evaluation.

\section{Pretest using Synthetic Dataset}

\begin{figure}[th]
\vskip 0.2in
\begin{center}
\resizebox{\linewidth}{!}{
\begin{tikzpicture}

\begin{groupplot}[group style={group size=4 by 1}, ticks=none,
title style={font=\LARGE, align=center}]

\nextgroupplot[
tick pos=left,
title={\huge \textbf{Class 1} \\ Rounded rectangle \\ Equilateral triangle},
xmin=-0.5, xmax=127.5,
label style={yshift=1em, align=center, font=\huge\bfseries},
ylabel={Rectangle-Triangle \\ dataset},
y dir=reverse,
ymin=-0.5, ymax=127.5
]
\addplot graphics [includegraphics cmd=\pgfimage,xmin=-0.5, xmax=127.5, ymin=127.5, ymax=-0.5] {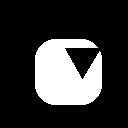};

\nextgroupplot[
tick pos=left,
title={\huge \textbf{Class 2} \\ Rounded rectangle \\ Right-angled triangle},
xmin=-0.5, xmax=127.5,
y dir=reverse,
ymin=-0.5, ymax=127.5
]
\addplot graphics [includegraphics cmd=\pgfimage,xmin=-0.5, xmax=127.5, ymin=127.5, ymax=-0.5] {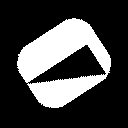};

\nextgroupplot[
tick pos=left,
title={\huge \textbf{Class 3} \\ Sharp rectangle \\ Equilateral triangle},
xmin=-0.5, xmax=127.5,
y dir=reverse,
ymin=-0.5, ymax=127.5
]
\addplot graphics [includegraphics cmd=\pgfimage,xmin=-0.5, xmax=127.5, ymin=127.5, ymax=-0.5] {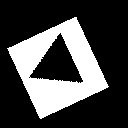};

\nextgroupplot[
tick pos=left,
title={\huge \textbf{Class 4} \\ Sharp rectangle \\ Right-angled triangle},
xmin=-0.5, xmax=127.5,
y dir=reverse,
ymin=-0.5, ymax=127.5
]
\addplot graphics [includegraphics cmd=\pgfimage,xmin=-0.5, xmax=127.5, ymin=127.5, ymax=-0.5] {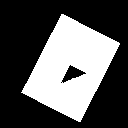};

\end{groupplot}

\end{tikzpicture}}
\resizebox{\linewidth}{!}{
\begin{tikzpicture}

\begin{groupplot}[group style={group size=4 by 1},ticks=none,
title style={font=\LARGE, align=center},
,/tikz/background rectangle/.style={draw=none}]
\nextgroupplot[
tick pos=left,
title={\huge \textbf{Class 1} \\ Circles only},
xmin=-0.5, xmax=127.5,
y dir=reverse,
label style={yshift=1em, align=center, font=\huge \bfseries},
ylabel={Circle-Square \\ dataset},
ymin=-0.5, ymax=127.5
]
\addplot graphics [includegraphics cmd=\pgfimage,xmin=-0.5, xmax=127.5, ymin=127.5, ymax=-0.5] {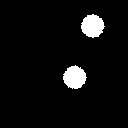};

\nextgroupplot[
tick pos=left,
title={Class 1},
xmin=-0.5, xmax=127.5,
title={\huge \textbf{Class 2} \\ Circle and Square},
y dir=reverse,
ymin=-0.5, ymax=127.5
]
\addplot graphics [includegraphics cmd=\pgfimage,xmin=-0.5, xmax=127.5, ymin=127.5, ymax=-0.5] {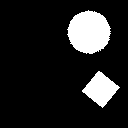};

\nextgroupplot[
tick pos=left,
title={\huge \textbf{Class 3} \\ Squares only},
xmin=-0.5, xmax=127.5,
y dir=reverse,
ymin=-0.5, ymax=127.5
]
\addplot graphics [includegraphics cmd=\pgfimage,xmin=-0.5, xmax=127.5, ymin=127.5, ymax=-0.5] {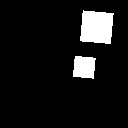};

\nextgroupplot[
tick pos=left,
title={},
axis line style={white},
xmin=-0.5, xmax=127.5,
y dir=reverse,
ymin=-0.5, ymax=127.5
]
\end{groupplot}

\end{tikzpicture}}
\caption{Representative examples of two synthetic datasets. All samples from the datasets were generated as binary images with 128x128 pixels. Size and position of individual objects is subject to randomization, whereas overall design is determined by code.}
\label{fig:toy-ds}
\end{center}
\end{figure}

We designed two synthetic datasets in order to assess performance of representation learning methods for classification tasks that require fine-grained image representations. This test bed enables to directly compare the linear classifiability of image representation from several methods. The two test scenarios are tailored to simulate data environments in which class discriminant information is either captured in a very fine-grained local feature or a holistic scene view -- see \autoref{fig:toy-ds}. On a general note, we are aware that there are no well-defined criteria to distinguish fine-grained  classification scenarios from other image classification tasks that require holistic image representations. In an attempt to probe these concepts, we designed two synthetic datasets, which are basis for the following experiment.

We find that the image representations provided through a contrastive task (i.e.~SimCLR) are well suited for the holistic scenario but they are not well suitable to exploit fine-grained details. In contrast, the image representations provided through a reconstruction task fall short for the holistic scenario, but they capture the fine-grained visual features. Image representations provided through ConRec are well suited for both scenarios -- see  \autoref{tab:synthetic}.

\subsection{Rationale of the Design}
\textbf{Rectangle-Triangle} -  We generate images that contain a rectangle with either right or round corners and a triangle, which is either equilateral or right-angled and placed inside the rectangle. By combining these two properties of each geometric object, we obtain four classes in total. In this case, multiple, disjoint low-level details in the images (angles in the triangle and shape of the rectangle corners) are important features for the classification. This favors approaches that generate fine-grained representations.

\vspace{2mm}
\noindent
\textbf{Circles-Square} - We generate images that contain either two circles (class 1), a square and a circle (class 2), or two squares (class 3).
For this task, it is not sufficient to represent the image exclusively through local information such as the as the edge of a single object. Instead, the representations must capture the identity and relation between the two objects and thus non-local information in order to be discriminative for the classification problem. This is what we refer to as \emph{holistic information} in an image.

\subsection{Results}
For each dataset, we generated a training and a test set with 1000/500 samples for the Rectangle-Triangle and 600/400 samples for the Circle-Square dataset. During pretraining, we evaluate the model's representations every 20 epochs and report the best results for each method after training for 1000 epochs in Table \ref{tab:synthetic}.

The results on the \emph{Rectangle-Triangle} dataset reveal that the model trained on the reconstruction task outperforms SimCLR and SimCLR-attention. Thus, the contrastive loss is not able to capture the fine-grained visual details in the representations. ConRec is able to capture these details and outperforms all other approaches. 

On the \emph{Circle-Square} dataset, SimCLR and SimCLR-attention clearly outperform the reconstruction task. Due to the holistic design of the dataset, this indicates that the reconstruction task is not well suited to create a corresponding representation of the image. It further suggests that the reconstruction task has its strength at encoding fine-grained features into the representations. In comparison, ConRec is capable to produce representations with holistic information and therefore almost reaches the performance of the two SimCLR variants. 

To conclude, the synthetic dataset results exemplify that ConRec is able to exploit both a holistic representation as well as fine-grained visual details of an image. Additional research is needed to assess the importance of individual reconstruction losses as our current implementation, the MSE, is known to have limitations \cite{pathak2016context}.

\begin{table}[h]
\centering
\caption{Linear evaluation accuracy on two synthetic datasets. The Rectangle-Triangle dataset requires local features and the Circles-Squares a holistic view on the image for good classification performance.}
\vspace{0.1in}
\begin{tabular}{|c|P{2cm}|P{2cm}|}
\hline
Method/Dataset  & Rectangle-Triangle & Circles-Squares  \\
\hline
SimCLR & 85.6   &  99.50    \\
SimCLR-attention  &  90.4   & 99.50 \\ 
Reconstruction   & 91.8 &  86.47 \\
ConRec with GA Pool   &   93.4  & 99.50 \\
ConRec   &   96.4  & 98.75 \\
\hline
\end{tabular}
\label{tab:synthetic}
\end{table}

\section{Benchmark Classification Tasks}
\label{sec:benchmark}

\begin{figure*}[b]
\begin{center}
\resizebox{\textwidth}{!}{\input{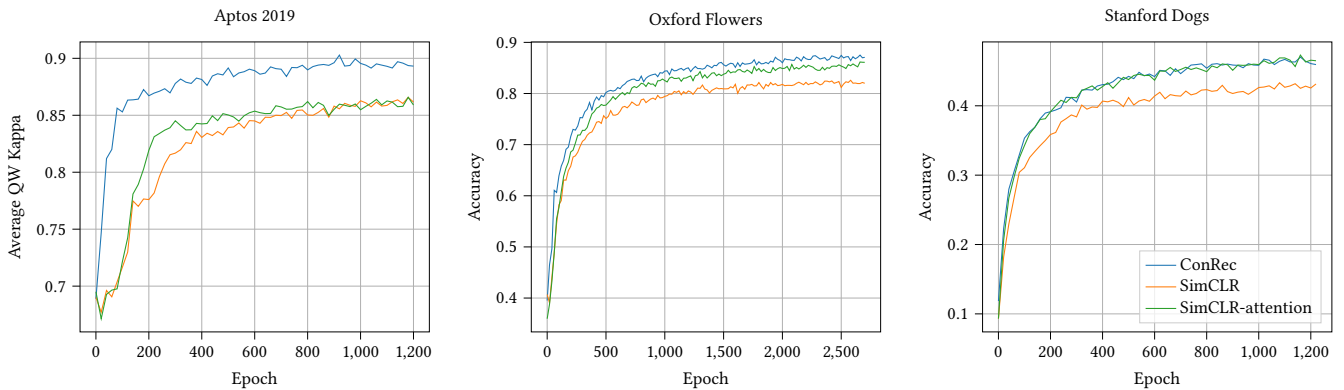}}
\caption{Linear evaluation performance during pretraining using three selected benchmark datasets. The models were evaluated every 20 epochs by fitting a logistic regression classifier on the image representations of the training data and then reporting classification metrics on fixed test data.}
\label{fig:pretraining-linear}
\end{center}
\end{figure*}

We evaluate our method on two publicly available fine-grained classification tasks and one medical dataset following our evaluation protocol described in Section \ref{sec:evaluation-protocol}.
We pretrain until convergence of the linear evaluation performance. This results in 1200 / 2700 / 1200 training epochs for the Aptos 2019 / Oxford Flowers / Stanford Dogs dataset respectively.
The evaluation performance during pretraining is depicted in Figure \ref{fig:pretraining-linear}. Final evaluation results with various training subsets are shown in Figure \ref{fig:benchmark}. 
Furthermore, we compare our linear evaluation results with other baselines in Table \ref{tab:benchmark}. We evaluate an ImageNet pretrained DenseNet121 with logistic regression in the same fashion as our pretrained models. Furthermore, we trained our U-net model (with standard global average pooling) and a DenseNet121 from scratch and report classification accuracy on the test set. For the linear evaluation results in Table \ref{tab:benchmark}, we additionally use crop augmentations when training the linear classifier which results in improved classification performance for the Stanford Dogs and Oxford Flowers dataset.

Comparing the results of models with self-supervised pretraining and a frozen encoder to models trained with a random initialization, the general benefit of using self-supervised pretraining becomes apparent. However, using the ConRec framework achieves the best results across all datasets except Stanford Dogs where it is en par with SimCLR-attention.

We explicitly report the linear performance results only and did not do any finetuning of the encoder, since the impact of self-supervised pretraining can be best evaluated in the linear evaluation regime \cite{chen2020simple}.

\begin{table*}[h]
\centering
\caption{Linear evaluation results and respective baselines. For the Aptos 2019 dataset, the evaluation metric is the average quadratic weighted kappa after performing 5-fold cross validation on the training set.
For all other datasets, categorical accuracy on the test set is reported. ImageNet results in parenthesis indicate flaws in the evaluation as the datasets were included in supervised ImageNet-pretraining.}
\vspace{0.1in}
\begin{tabular}{|l|c|c|c|c|c|}
\hline
Model & Frozen Encoder & Aptos 2019 & Oxford Flowers & Stanford Dogs & \#Parameters  \\
\hline
SimCLR U-net & \checkmark &  85.72  & 86.01 & 43.96 & 4.693M  \\
SimCLR Attention U-net & \checkmark  &   86.06  & 88.37  & 50.31 & 4.867M \\ 
ConRec U-net   &  \checkmark &  89.44  & 90.29 & 49.57 & 4.867M \\
DenseNet121 (ImageNet Init) & \checkmark &    86.70  & (92.97) & (88.07) & 8.062M \\
\hline
U-net (Random Init) & & 82.11  &    81.54  & 55.2 & 4.693M  \\
DenseNet121 (Random Init)  & & 64.53  &  82.03    & 57.63 & 8.062M \\
\hline
\end{tabular}
\label{tab:benchmark}
\end{table*}

\subsection{Oxford Flowers} 
The Oxford Flowers dataset consists of 102 classes of flowers. Similar to \citet{chen2020simple}, we use the official train and validation split combined as the training set which contains 20 images per class resulting in 2040 total images. For evaluation, we use the official test split with 6,149 samples.

The results show that the attention mechanism helps to improve the representations of the SimCLR framework across all subsets. The reconstruction task additionally improves the representations by a significant margin. Especially in low data regimes, the performance gain increases.

\subsection{Diabetic Retinopathy Classification (Aptos 2019)}
We also evaluate all models on a medical dataset where fine grained details are often important for classification performance. We use the Aptos 2019 Kaggle Challenge \cite{aptos2019} which also has been studied in the context of self-supervised learning by \citet{taleb20203d}. As the labels of the test set are not public, we follow their approach and evaluate our models with 5-fold cross validation on the training set. The evaluation metric in this case is the average weighted quadratic kappa. Our results show that the attention mechanism only moderately increases the performance in this case. However, the addition of the reconstruction task results in a significant performance gain. 

\citet{welikala2014automated, welikala2015genetic} perform extensive feature engineering for diabetic retinopathy detection. They show that the morphology of the local retinal vasculature is important for classification performance. We suppose that this are local features that can be better captured by the reconstruction task and therefore result in an increased linear performance.

\begin{figure}[h]
\begin{center}
\resizebox{.8\linewidth}{!}{
\begin{tikzpicture}

\definecolor{color0}{rgb}{0.12156862745098,0.466666666666667,0.705882352941177}
\definecolor{color1}{rgb}{1,0.498039215686275,0.0549019607843137}
\definecolor{color2}{rgb}{0.172549019607843,0.627450980392157,0.172549019607843}
\definecolor{color3}{rgb}{0.83921568627451,0.152941176470588,0.156862745098039}
\definecolor{color4}{rgb}{0.580392156862745,0.403921568627451,0.741176470588235}
\definecolor{color5}{rgb}{0,0,0}

\definecolor{black}{rgb}{0,0,0}
\definecolor{red}{rgb}{1,0,0}
\definecolor{green}{rgb}{0,0.5,0}
\definecolor{blue}{rgb}{0,0,1}

\begin{axis}[
legend cell align={left},
legend style={fill opacity=0.8, draw opacity=1, text opacity=1, at={(0.97,0.03)}, anchor=south east, draw=white!80!black},
tick align=outside,
tick pos=left,
x grid style={white!69.0196078431373!black},
xlabel={Epoch},
xmajorgrids,
xmin=-60, xmax=1260,
xtick style={color=black},
y grid style={white!69.0196078431373!black},
ylabel={Avg QW Kappa},
ymajorgrids,
ymin=0.646016564965248, ymax=0.91,
ytick style={color=black}
]
\addplot [semithick, green, dashed]
table {%
0 0.694925963878632
20 0.68043851852417
40 0.656485497951508
60 0.700958371162415
80 0.708566188812256
100 0.709946930408478
120 0.689910054206848
140 0.703423976898193
160 0.713230907917023
180 0.723460674285889
200 0.718910992145538
220 0.720954537391663
240 0.717015147209167
260 0.715399205684662
280 0.762474834918976
300 0.77431982755661
320 0.778525471687317
340 0.778008341789246
360 0.782738506793976
380 0.793093144893646
400 0.780097484588623
420 0.787195920944214
440 0.79036808013916
460 0.782801330089569
480 0.795117974281311
500 0.803129851818085
520 0.799978137016296
540 0.810862898826599
560 0.81772118806839
580 0.81943541765213
600 0.827605605125427
620 0.810476779937744
640 0.819062113761902
660 0.82244873046875
680 0.822974860668182
700 0.82810389995575
720 0.824203968048096
740 0.834000945091248
760 0.831697821617126
780 0.837269484996796
800 0.833308219909668
820 0.831284701824188
840 0.827529430389404
860 0.828751862049103
880 0.829247653484344
900 0.824918150901794
920 0.835500538349152
940 0.837157547473907
960 0.836447060108185
980 0.836326956748962
1000 0.837376475334167
1020 0.847365379333496
1040 0.842476546764374
1060 0.836439967155457
1080 0.841775059700012
1100 0.836594939231873
1120 0.840284645557404
1140 0.843694984912872
1160 0.831133008003235
1180 0.847938358783722
1200 0.841870307922363
};
\addlegendentry{SimCLR-4}
\addplot [semithick, blue, dashed]
table {%
0 0.707291841506958
20 0.676405131816864
40 0.711518526077271
60 0.695243358612061
80 0.687965512275696
100 0.700663328170776
120 0.71676504611969
140 0.73091197013855
160 0.762405097484589
180 0.75994884967804
200 0.767187833786011
220 0.782597839832306
240 0.781177222728729
260 0.795670986175537
280 0.798826575279236
300 0.801017165184021
320 0.800700962543488
340 0.804861426353455
360 0.811547636985779
380 0.813488006591797
400 0.817228615283966
420 0.819961249828339
440 0.8258056640625
460 0.824478983879089
480 0.833595156669617
500 0.827329277992249
520 0.835515201091766
540 0.833319962024689
560 0.840908646583557
580 0.839729428291321
600 0.844394505023956
620 0.846512615680695
640 0.850178360939026
660 0.841487228870392
680 0.848875045776367
700 0.851594924926758
720 0.849978744983673
740 0.846677958965302
760 0.854158282279968
780 0.852597415447235
800 0.849584877490997
820 0.852958023548126
840 0.852224946022034
860 0.855997383594513
880 0.859608471393585
900 0.850557804107666
920 0.858350396156311
940 0.857729077339172
960 0.854920089244843
980 0.861899495124817
1000 0.8649982213974
1020 0.859105408191681
1040 0.864352703094482
1060 0.861186504364014
1080 0.865565180778503
1100 0.855126976966858
1120 0.861933588981628
1140 0.857265949249268
1160 0.864610850811005
1180 0.863643288612366
1200 0.865864157676697
};
\addlegendentry{SimCLR-8}
\addplot [semithick, black, dashed]
table {%
0 0.690142512321472
20 0.67711067199707
40 0.696204304695129
60 0.690644383430481
80 0.703706324100494
100 0.716595709323883
120 0.729849934577942
140 0.774787545204163
160 0.770041346549988
180 0.77650660276413
200 0.775989830493927
220 0.78174215555191
240 0.796217203140259
260 0.80737030506134
280 0.815349876880646
300 0.816765010356903
320 0.819726169109344
340 0.825992405414581
360 0.825181484222412
380 0.835848927497864
400 0.830576241016388
420 0.834224343299866
440 0.832187831401825
460 0.835722744464874
480 0.832814872264862
500 0.83922415971756
520 0.839735984802246
540 0.843237698078156
560 0.838695347309113
580 0.845292866230011
600 0.845041275024414
620 0.842973530292511
640 0.848262190818787
660 0.848137259483337
680 0.849999785423279
700 0.849708735942841
720 0.852534890174866
740 0.847292900085449
760 0.854269802570343
780 0.854622960090637
800 0.850254237651825
820 0.8499356508255
840 0.852521061897278
860 0.855960845947266
880 0.848337292671204
900 0.858055472373962
920 0.855653882026672
940 0.860465824604034
960 0.858788192272186
980 0.857830047607422
1000 0.862743973731995
1020 0.860875487327576
1040 0.857423007488251
1060 0.861551105976105
1080 0.858068764209747
1100 0.858677685260773
1120 0.861825287342072
1140 0.863527119159698
1160 0.860361099243164
1180 0.865738272666931
1200 0.861584782600403
};
\addlegendentry{SimCLR-16}
\addplot [semithick, green]
table {%
0 0.708689749240875
20 0.681882619857788
40 0.691608071327209
60 0.676570534706116
80 0.662247657775879
100 0.712275445461273
120 0.701318621635437
140 0.721673488616943
160 0.761890769004822
180 0.772146344184875
200 0.774272561073303
220 0.780470490455627
240 0.793232083320618
260 0.793541550636292
280 0.795160830020905
300 0.807807624340057
320 0.799642622470856
340 0.807471573352814
360 0.805206418037415
380 0.81689989566803
400 0.817613244056702
420 0.815185546875
440 0.823184132575989
460 0.822592079639435
480 0.829225659370422
500 0.829070091247559
520 0.829213798046112
540 0.830410659313202
560 0.834017634391785
580 0.833906531333923
600 0.840705692768097
620 0.834132194519043
640 0.838354408740997
660 0.845714867115021
680 0.844710469245911
700 0.846428751945496
720 0.84456330537796
740 0.845253109931946
760 0.850917816162109
780 0.848988831043243
800 0.846352756023407
820 0.848014533519745
840 0.842738330364227
860 0.844112396240234
880 0.844937205314636
900 0.850817799568176
920 0.854848086833954
940 0.853721797466278
960 0.852047920227051
980 0.853758633136749
1000 0.854718387126923
1020 0.856199860572815
1040 0.861072063446045
1060 0.852332293987274
1080 0.852114200592041
1100 0.855381309986115
1120 0.857389152050018
1140 0.847915768623352
1160 0.854465365409851
1180 0.854192554950714
1200 0.854362368583679
};
\addlegendentry{SimCLR-32}
\addplot [semithick, blue]
table {%
0 0.710167407989502
20 0.705309748649597
40 0.712561011314392
60 0.714887261390686
80 0.71272200345993
100 0.700165867805481
120 0.720249772071838
140 0.724635899066925
160 0.738361954689026
180 0.760347723960876
200 0.778431057929993
220 0.787556529045105
240 0.799043953418732
260 0.801139652729034
280 0.821378409862518
300 0.820678353309631
320 0.825547397136688
340 0.828301310539246
360 0.831338584423065
380 0.838371455669403
400 0.838782489299774
420 0.840567111968994
440 0.845060348510742
460 0.84672087430954
480 0.847418963909149
500 0.842443585395813
520 0.849094271659851
540 0.848418414592743
560 0.852434992790222
580 0.845871150493622
600 0.852284729480743
620 0.850012481212616
640 0.853114902973175
660 0.851569652557373
680 0.851668536663055
700 0.855442345142365
720 0.854494452476501
740 0.851527988910675
760 0.859162330627441
780 0.857131123542786
800 0.857473492622375
820 0.856530964374542
840 0.855121910572052
860 0.854866623878479
880 0.857444763183594
900 0.854177832603455
920 0.853039383888245
940 0.858547031879425
960 0.858532249927521
980 0.854424774646759
1000 0.856927871704102
1020 0.856339931488037
1040 0.860449433326721
1060 0.85979700088501
1080 0.859830558300018
1100 0.855444312095642
1120 0.861354470252991
1140 0.858001112937927
1160 0.862101912498474
1180 0.860113739967346
1200 0.860633075237274
};
\addlegendentry{SimCLR-64}

\addplot [thick, black]
table {%
0 0.688889145851135
20 0.746764183044434
40 0.81200110912323
60 0.819998443126678
80 0.856233894824982
100 0.852951526641846
120 0.863316178321838
140 0.863563239574432
160 0.864147186279297
180 0.872574985027313
200 0.867154240608215
220 0.86946713924408
240 0.871098518371582
260 0.873313903808594
280 0.869676887989044
300 0.877940654754639
320 0.881804823875427
340 0.879026591777802
360 0.877949893474579
380 0.882666230201721
400 0.881403923034668
420 0.87623530626297
440 0.884481549263
460 0.886448681354523
480 0.885418772697449
500 0.891518592834473
520 0.883826434612274
540 0.887137591838837
560 0.888239502906799
580 0.890471756458282
600 0.889122664928436
620 0.88604199886322
640 0.887061953544617
660 0.892474055290222
680 0.890936851501465
700 0.890419185161591
720 0.884072303771973
740 0.891949474811554
760 0.891757667064667
780 0.893906116485596
800 0.889837622642517
820 0.892746090888977
840 0.894036114215851
860 0.894603610038757
880 0.893812775611877
900 0.896148085594177
920 0.902819812297821
940 0.893186569213867
960 0.893727004528046
980 0.899458289146423
1000 0.895553588867188
1020 0.894009590148926
1040 0.891426682472229
1060 0.89512026309967
1080 0.89403235912323
1100 0.89296692609787
1120 0.891500174999237
1140 0.897021412849426
1160 0.895771026611328
1180 0.893633663654327
1200 0.893145561218262
};
\addlegendentry{ConRec-16}
\end{axis}

\end{tikzpicture}}
\caption{SimCLR pretraining performance on our U-net encoder using standard global average pooling on the Aptos2019 dataset for batch sizes 4, 8, 16, 32, 64 in comparison to ConRec with batch size 16.}
\label{fig:batch-size}
\end{center}
\end{figure}
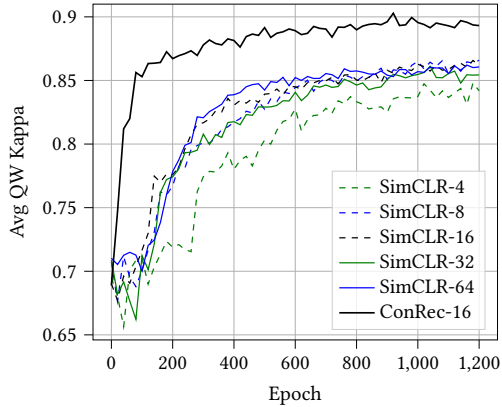

We also investigated, if the reduced batch size is a limiting factor of our SimCLR implementation. Therefore, we pretrained the SimCLR implementation with batch sizes 4, 8, 16, 32 and 64 and show the average linear performance during pretraining in \autoref{fig:batch-size}. The results show that although the performance significantly increases from batch size 4 to 8, an additional batch size increase does not result in a significant performance gain.  

Comparing our results to \citet{taleb20203d}, we achieve significantly higher quadratic weighted kappa score while only using the training data from the Kaggle Challenge during pretraining and performing linear evaluation on the representations as opposed to finetuning the whole model. 

Furthermore, ConRec’s representations also outperform the representations of an ImageNet-pretrained DenseNet121. This highlights the benefits of in-domain pretraining rather than transferring from pretrained models that were trained on large out-of-domain datasets. This confirms both the initial observation by \citet{he2019rethinking} stating that ImageNet-pretraining does not necessarily lead to better performance in other tasks and also \citet{ke2021chextransfer}, which shows that ImageNet performance does not correlate with performance on fine-grained classification of medical images (specifically CheXpert in this example). 

\subsection{Stanford Dogs} 
The Stanford Dogs dataset contains images of 120 breeds of dogs and has been built using images and annotation from ImageNet. The official train set contains 100 images per class resulting in a train set with 12000 images. For evaluation, we use the official test split with 8580 samples.

The results show that the attention mechanism improves the representations for this dataset. However, in contrast to the two other datasets, the addition of the reconstruction task does not result in any performance improvement and ConRec performs equally to SimCLR-attention.

It should be noted that all self-supervised representations are also significantly outperformed by the ImageNet-pretrained DenseNet121. This margin can be explained by the fact that this dataset is a subset of the ImageNet dataset and therefore, the DenseNet121 was exposed to further dog images and true label information during pretraining.

\section{Future Work}
Due to computational limitations, we were unable to pretrain our model on the ImageNet dataset.
This would allow us to compare our representations with other recent advancements in the self-supervised learning space and could help to further study the transferability of the learned representations. Furthermore, adding a decoder increases the memory need. Therefore, a momentum encoder combined with memory bank techniques as presented in MoCo \cite{he2020momentum} could potentially improve our method. The addition of an adversarial loss which has shown to produce sharper reconstruction predictions \cite{pathak2016context, isola2017image} could also help to capture even more fine-grained features in the image. \citet{chen2020big} showed that intermediate layers in the projection head may build better representations in low data regimes. In our experiments, the representations at the encoder output outperformed the ones in the projection head. 
Future work should investigate the design of the projection head in combination with the attention mechanism and a reconstruction task.
We focused on a U-net architecture \cite{ronneberger2015u} for pretraining our models. Further research should also investigate if our self-supervised model pretraining can be beneficial for segmentation tasks as it was shown for other reconstruction-based approaches \cite{pathak2016context}.

\section{Related Work}
\label{related_work}
Self-supervised learning methods aim to construct semantically meaningful representations in the absence of labels or any semantic annotations.  Self-supervised learning has been widely studied for several domains such as image \cite{misra2020self}, audio \cite{tagliasacchi2019self} or multiple modalities \cite{owens2018audio, taleb2019multimodal, radford2learning}. 
Three types of learning tasks are commonly used for self-supervised learning in computer vision tasks. Models trained with handcrafted pretext tasks predict properties of an image such as colorization \cite{zhang2016colorful}, rotation \cite{gidaris2018unsupervised}, or patch relatedness \cite{doersch2015unsupervised}.   

Self-reconstruction tasks let the model learn to reconstruct an artificially corrupted input. 
A simple but successful self-reconstruction approach is the in-painting task \cite{pathak2016context} where the model generates the content of an arbitrary image patch conditioned on its surrounding.
In Model Genesis \cite{zhou2019models} this concept was extended to 3D medical images and the model restores the complete image while parts of the input were occluded. In the Parts2Whole approach \cite{feng2020parts2whole}, the model learns part-whole semantics by reconstructing a whole (i.e.~a 3D image) from its parts (i.e.~several 3D crops). Due to the generative nature of the task, self-reconstruction models are commonly implemented as encoder-decoder architectures, which make them suitable for transfer learning in segmentation problems \cite{pathak2016context}.

For contrastive learning tasks, the model aims to generate a representation that maximizes the similarity between positive pairs while minimizing the similarity between negative pairs. In visual representation learning, positive/negative pairs are commonly generated by sampling two patches from the same/different image. Multiple approaches for contrastive image representation have recently been proven successful: SimCLR \cite{chen2020simple, chen2020big} provides a simple learning framework that allows to sample negative pairs within each mini batch. Therefore, SimCLR benefits from larger batch sizes and it can be trained without a memory bank of negative pairs. Having a similar learning target, Momentum Contrast \cite{he2020momentum, chen2020improved} can be trained with smaller batch sizes as it keeps a buffer of negatives for loss calculation.

Other recent self-supervised learning approaches also omit extensive negative mining. SwAV \cite{caron2020unsupervised} uses a clustering-based approach enforcing consistency between cluster assignments produced for different augmentations of the same image. BYOL \cite{grill2020bootstrap} trains an online network to predict the target network's representation of the same image under a differently augmented view while updating the target network with a momentum term. Image GPT \cite{chen2020generative} learns image representations by completing masked images in an autoregressive fashion with a transformer model.

Instead of contrasting two augmentations from the same image, CLIP \cite{radford2learning} learns to correctly align images to their corresponding captions.
After training on 400 million (image, text) pairs, the model achieves SOTA image representations and enables zero-shot transfer of the model to downstream tasks.

Attention mechanisms were used in other approaches to tackle fine-grained classification:
\citet{han2018attribute} introduce an Attribute-Aware Attention Model which uses additional attribute information about the image combined with category label information to build more discriminate feature representations. \citet{chang2020devil} utilizes a channel-wise attention mechanism which forces all feature channels belonging to the same class to be discriminative across the network. \citet{chen2019destruction} shuffles regions of the input image and trains a region alignment network to restore the original spatial layout.

\section{Conclusion}

We showcased that the SimCLR framework has shortcomings to capture fine-grained features in its representations. To address this issue, we introduced an attention pooling mechanism and a reconstruction task. On two fine-grained classification datasets and one medical dataset, the attention pooling mechanism improves self-supervised learning performance significantly. The addition of the reconstruction task yields a further performance increase for diabetic retinopathy classification and on the Oxford Flowers dataset.

\bibliographystyle{ACM-Reference-Format}
\bibliography{main.bib}


\end{document}